\documentclass[10pt,twocolumn,letterpaper]{article}

\usepackage[utf8]{inputenc}
\usepackage[T1]{fontenc}

\usepackage{amsmath, amssymb, bm, cases, multirow, booktabs, array, tabularx}
\usepackage{float, placeins}
\usepackage{arydshln}

\usepackage[dvipsnames, table]{xcolor}  
\usepackage{graphicx}
\usepackage{svg}

\usepackage{cuted}
\usepackage{CJKutf8}
\usepackage{caption}
\usepackage{algorithm}
\usepackage{algorithmic}
\usepackage{ulem}
\usepackage{pifont}

\usepackage{cvpr}              

\usepackage{hyperref}

\usepackage{cvpr}              

\usepackage{hyperref}

\usepackage{pgfplots}
\usepackage{adjustbox}

\def\paperID{1885} 
\def\confName{CVPR}
\def\confYear{2025}

\title{
IDProtector: An Adversarial Noise Encoder to Protect Against ID-Preserving Image Generation
}

\author{
Yiren Song\thanks{Equal contribution.} \quad Pei Yang\footnotemark[1] \quad Hai Ci\footnotemark[2] \quad Mike Zheng Shou\thanks{Corresponding author.} \\
Show Lab, National University of Singapore \\
}

\begin{document}
\maketitle

\begin{strip}
\centering
    \includegraphics[width=\linewidth]{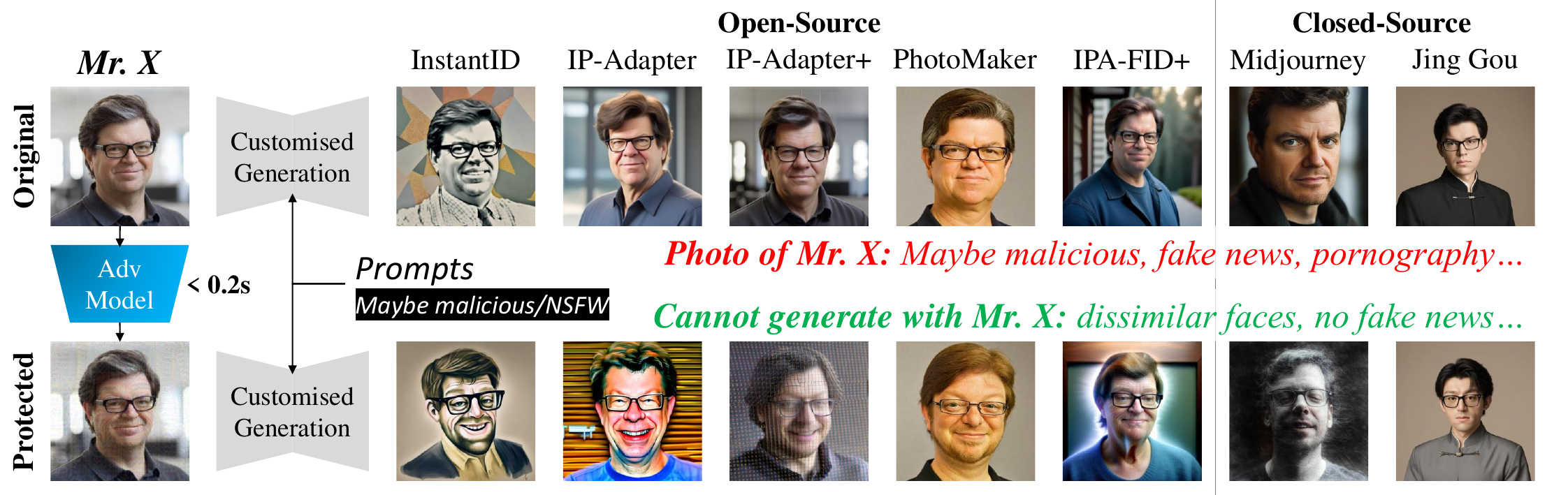}
    \vspace{-7pt}
    \vspace{-0.1cm}
    \captionof{figure}{Using only a single reference image, Encoder-based ID-preserving methods can generate realistic portraits, posing serious risks for malicious uses such as generation of malicious images. To prevent unauthorized ID-preserving generation, we propose IDProtector, which adds small perturbations to images. The protected images could disrupt ID-preserving generation by misleading the customization models to generate a dissimilar face, thereby achieving protection.}
    \label{fig:teaser}
\end{strip}

\begin{abstract}
Recently, zero-shot methods like InstantID have revolutionized identity-preserving generation. Unlike multi-image finetuning approaches such as DreamBooth, these zero-shot methods leverage powerful facial encoders to extract identity information from a single portrait photo, enabling efficient identity-preserving generation through a single inference pass. 
However, this convenience introduces new threats to the facial identity protection.
This paper aims to safeguard portrait photos from unauthorized encoder-based customization. 
We introduce IDProtector, an adversarial noise encoder that applies imperceptible adversarial noise to portrait photos in a single forward pass. 
Our approach offers universal protection for portraits against multiple state-of-the-art encoder-based methods, including InstantID, IP-Adapter, and PhotoMaker, while ensuring robustness to common image transformations such as JPEG compression, resizing, and affine transformations.
Experiments across diverse portrait datasets and generative models reveal that IDProtector generalizes effectively to unseen data and even closed-source proprietary models. Project page: \href{https://github.com/showlab/IDProtector}{https://github.com/showlab/IDProtector}.

\end{abstract}




\section{Introduction}
\label{sec:intro}

Recent advancements in diffusion models have opened new avenues for high-fidelity ID-preserving generation, widely used in applications such as personalized avatars. 
These methods can be categorized into tuning-based and encoder-based approaches.
While tuning-based methods like Dreambooth\cite{dreambooth} and LoRA\cite{lora} are effective, they require finetuning across multiple photos, which can be inconvenient. 
Encoder-based techniques, such as InstantID\cite{instantid} and IP-Adapter\cite{ipa}, leveraging a face encoder to extract facial identity features in a zero-shot way, eliminating the need for finetuning and significantly simplifying the customization process. 
However, these advancements also intensify threats to facial privacy and security, making the protection of personal identity information in portrait photos increasingly critical.
In previous research, Anti-DreamBooth\cite{antidreambooth} utilized PGD\cite{pgd} optimization against finetuning-based methods like LoRA and Dreambooth. Yet, there remains a gap in the literature regarding attacks on encoder-based ID-preserving generation.

Encoder-based ID preservation technology evolves from domain-agnostic encoders to domain-specific ones. 
Initially, methods like Face0\cite{face0} and IP-Adapter\cite{ipa} used a general CLIP\cite{clip} image encoder to extract facial features, yet they struggled to maintain the identity fidelity. 
IP-Adapter-Plus\cite{ipa} addressed this issue by utilizing patch features before the projection layer. InstantID\cite{instantid} introduced the use of a pre-trained Arcface\cite{arcface} encoder, significantly enhancing identity consistency and influencing subsequent works like Photomaker~\cite{photomaker} and PULID~\cite{pulid}. 
A natural progression is to attack both the CLIP image encoder and Arcface encoder simultaneously, targeting all current Encoder-based ID preservation technologies.

There are four major challenges in ID protection: 
\textbf{1. Universality.} Given the numerous existing encoder-based ID-preserving methods, we believe that universal protection is necessary, as attackers can easily switch 
to other methods to bypass protection. 
\textbf{2. Efficiency.} 
Previous work typically relies on the Projected Gradient Descent (PGD)~\cite{pgd} to generate adversarial noise. However, PGD requires per-image optimization, taking several minutes per image, which significantly limits its practicality for protecting large volumes of portrait photos on social media in real world.
\textbf{3. Robustness.} The adversarial noise must be robust enough to withstand common image transformations, such as cropping, resizing, and JPEG compression, while also countering the inherent pre-processing operations of identity-preserving methods, such as face alignment.
\textbf{4. Imperceptibility.} The adversarial noise added for ID protection should be imperceptible to human eyes.

In response to these challenges, this paper presents \textbf{IDProtector}, the first feed-forward method to protect portrait photos from state-of-the-art encoder-based identity preservation techniques. Our approach incorporates several key innovations. For efficiency, we propose a ViT-based encoder that directly predicts adversarial noise for a given target portrait. To ensure universality, we carefully examine the attack surfaces and define a composite loss function tailored to four mainstream encoder-based generation methods, including InstantID~\cite{instantid}, IP-Adapter~\cite{ipa}, IP-Adapter Plus~\cite{ipa}, and PhotoMaker~\cite{photomaker}. Additionally, we employ an L1-based image regularization to minimize the visual impact of adversarial noise on image quality.
For robustness against common transformations, a typical approach involves a combination of post-hoc differentiable augmentations like differentiable JPEG and cropping during training. However, we find that this method introduces a substantial learning burden. To address this, we propose a surrogate method that leverages the face preprocessing pipeline of existing identity-preserving methods. Specifically, we introduce small random noise into the affine transformation matrix during face alignment stage to achieve data augmentation. 
This not only enhances the robustness of adversarial noise against common image transformations but also enables it to withstand image preprocessing operations during identity-preserving generation.

We train and extensively evaluate IDProtector on the CelebA dataset~\cite{celeba}. We further test it on unseen data VGG Face~\cite{vggface} and proprietary models like Midjourney~\cite{midjourney} and Jing Gou, demonstrating its strong generalization to unseen data and closed-source models. 
By advancing adversarial attack techniques from theoretical constructs to practically deployable solutions, IDProtector sets a new benchmark in facial privacy protection.
We summarize our main contributions as follows:
\begin{itemize}
    \item We introduce \textbf{IDProtector}, the first feed-forward method to protect portrait photos from SOTA encoder-based identity preservation techniques.
    \item We present several innovative techniques including a ViT-based noise encoder, a joint learning framework with carefully designed objectives, and affine-based data augmentation to simultaneously achieve the \textbf{EURI} standard (\textbf{E}fficiency, \textbf{U}niversality, \textbf{R}obustness, and \textbf{I}mperceptibility).
\end{itemize}

\section{Related Works}
\label{sec:related}

\subsection{Tuning-based Customized Generation}

With the rapid advancement of diffusion models, customized generation methods have begun to emerge.
Techniques like DreamBooth \cite{dreambooth}, Custom Diffusion \cite{customdiffusion}, Low-Rank Adaptation (LoRA) \cite{lora} and Textual Inversion \cite{textualinversion} facilitate the adaptation of text-to-image diffusion models through fine-tuning strategies that range from adjusting all parameters to optimizing specific word vectors. Despite their impressive effects, these methods require extensive computational resources and time due to their need for multi-image fine-tuning. This can make them impractical for real-time or large-scale applications.



\subsection{Encoder-based ID-Preserving Generation}


Encoder-based Customized generation methods\cite{ipa,face0,ssr,fast} utilize pre-trained encoders to extract appearance feature, facilitating zero-shot generation in real-time. These methods evolved from domain-agnostic encoders to domain-specific encoders. Initially, CLIP\cite{clip} image encoder was used to extract facial features, such as with Face0\cite{face0} and IP-Adapter\cite{ipa}; however, their maintenance of facial detail fidelity was not satisfactory. IP-Adapter-Plus addressed this issue by utilizing patch features before the projection layer. InstantID\cite{instantid} was the first to propose using a pre-trained Arcface\cite{arcface} Encoder as a condition, significantly enhancing identity consistency and influencing subsequent works like Photomaker\cite{photomaker} and PULID\cite{pulid}. StoryMaker\cite{storymaker} employs both the Arcface image encoder and the CLIP image encoder to encode facial features and broader regions, respectively, preserving not only facial identities but also clothing, hairstyles, and body features. Additionally, encoder-based ID preservation technology is widely applied in multi-human image personalization\cite{uniportrait, moa} and video tasks\cite{idanimator}.



\begin{figure*}[ht]
    \centering
    \includegraphics[width=1\linewidth]{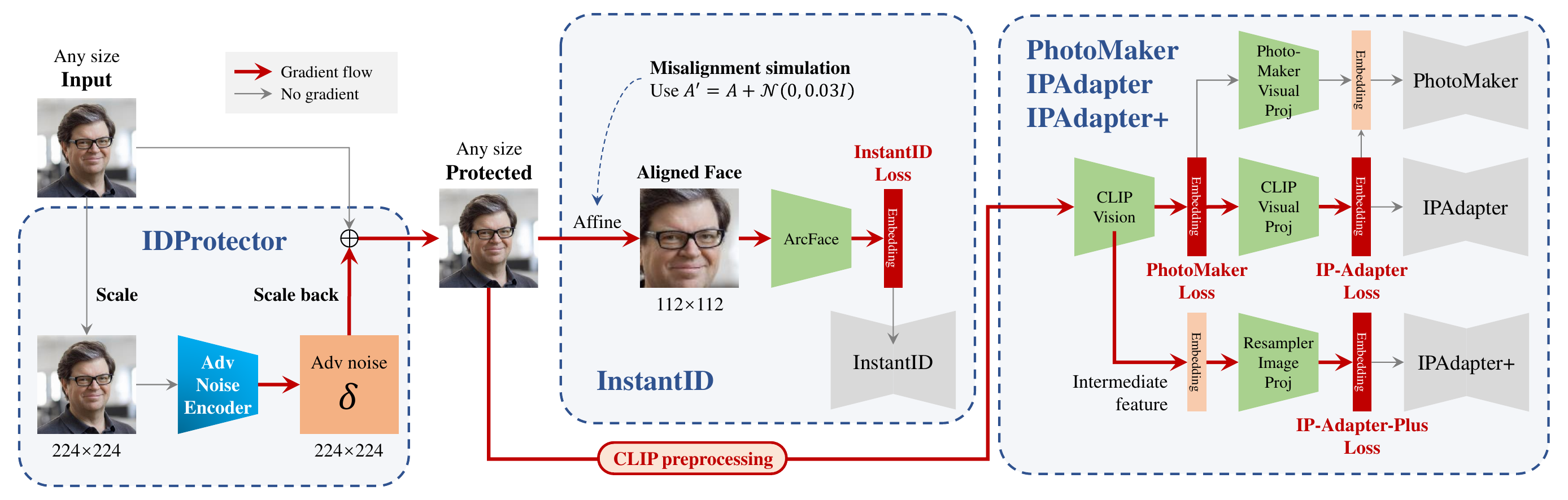}
    \caption{Overall schematics of our method. Our method's key design includes the noise encoder, loss functions, and the gradient optimization path that allows backpropagation. During the protection phase, an image is first resized to 224$\times$224 and sent to an adversarial noise encoder model. The model outputs a perturbation that can be resized back and added to the image, achieving an adversarial attack effect -- the protected image prevents subsequent pipelines from correctly extracting facial feature information, thereby hindering face similarity in face generation tasks. This achieves the goal of ID protection.}
    \label{fig:architecture}
\end{figure*}

\subsection{Defensive Perturbations for Diffusion Models.}

The focus on safeguarding personal artworks and facial images from unauthorized usage in the fine-tuning of diffusion models has led to the development of subtle yet effective perturbations. As summarized in Tab. \ref{tab:xxx}, distinct approaches have been tailored to address this issue: Glaze \citep{glaze} strategically adjusts the feature-space distances to misdirect the style learning process in Stable Diffusion, effectively preventing style appropriation. Conversely, AdvDM \citep{advdm} implements adversarial disruptions by maximizing the Mean Squared Error during the optimization phase to shield personal visuals. Meanwhile, Anti-DreamBooth \citep{antidreambooth} integrates adversarial interference directly within the DreamBooth fine-tuning cycle, applying a complex optimization strategy to formulate defensive perturbations. 
Further investigations \citep{simac, duaw, ace, song2024anti} into adversarial protection have demonstrated their effectiveness in obstructing finetuning-based customization methods such as DreamBooth \citep{dreambooth}, LoRA \citep{lora}, and textual inversion \citep{textualinversion}. However, how to protect personal identity from the recently popular encoder-based methods remains unexplored. This paper endeavors to bridge the current gaps by proposing advanced methodologies to shield images from encoder-based exploitations, enhancing the security measures for personal digital assets. 

\begin{table}[ht]
    \centering
    \caption{Overview of encoder-based methods for ID-preserving generation.}
    \setlength{\tabcolsep}{0pt}
    \resizebox{\linewidth}{!}{ 
    \begin{tabular}{c c} 
    \toprule
    \textbf{Encoder} & \textbf{ID-Preserving Methods} \\ 
    \hline
    CLIP \cite{clip} & IP-Adapter \cite{ipa}, IP-Adapter Plus \cite{ipa}, PhotoMaker \cite{photomaker}, Face0 \cite{face0} \\ 
    ArcFace \cite{arcface} & InstantID \cite{instantid}, PhotoMaker \cite{photomaker}, PuLID \cite{pulid}, IP-Adapter-FaceID \cite{ipa} \\ 
    CLIP \& ArcFace  & IDAdapter \cite{idadapter},  Storymaker \cite{storymaker} \\ 
    \bottomrule
    \end{tabular}
    }
    \label{tab:xxx}
\end{table}

\section{Problem Definition}
We consider two participants in a portrait identity protection scenario.: "image protector" Alice and "image exploiter" Bob. 
Bob seeks to use an encoder-based, identity-preserving approach to alter the portrait photos Alice shares on social media for malicious purposes.
While Alice aims to protect her portrait by adding adversarial noise to thwart Bob's methods, inducing distortions or significant decreases in ID similarity in the generated content. Specifically, we explain the workflow of the two parties as follows:

\textbf{1. Image Protector Alice:} The image protector aims to provide protection for images against exploitation by encoder-based ID-preserving methods. 
In this context, the chosen protection method involves adding imperceptible protective perturbations to the images, with the goal of providing protection while minimizing modifications to the original image. 
In real-world scenarios, image protectors encounter several challenges. For instance, implementing universal protection to guard against various encoder-based, identity-preserving methods is difficult. Additionally, the protection should be robust to natural distortions such as resizing, JPEG compression, and blurring.

\textbf{2. Image Exploiter Bob:} Bob aims to use portrait photos for ID-preserving customization. Bob can choose any open-source or proprietary encoder-based models, which poses significant challenges on the universality of the protection method.

\textbf{Objective of the Study:} Given an image $x$ containing a face portrait, our objective is to generate adversarial perturbations $\delta$ that protect against ID-preserving generation. We aim to maximize the difference in identity consistency between the unperturbed generated image $g(x)$ and the perturbed generated image $g(x+\delta)$, where $g$ is an ID-preserving generator. To unify the protection against different models, we can find a $\delta$ that simultaneously disrupts several feature spaces, lowering the sum of the similarity values. Thus, we train a noise encoder network $E_\theta$, solving
\begin{equation}
    \min_\theta S
    \left\{
        g_i(x), g_i(x+\delta)
    \right\}, \quad \delta = E_\theta(x)
\end{equation}
\noindent subject to the constraint that $\lVert \delta \rVert_\infty < \epsilon$. Here, $S$ denotes an identity similarity metric, which we evaluate using cosine similarity in feature spaces of ArcFace \cite{arcface} and CLIP \cite{clip}.

\section{Method}

The core of our method is an adversarial noise encoder model that can add adversarial noise to facial images for protection. We will first introduce the complete workflow in Sec. \ref{sec:overall_method}, followed by the design of the adversarial noise encoder in Sec. \ref{sec:adv_noise_encoder}. Sec. \ref{sec:IP_protection_obj} and \ref{sec:loss_functions} introduce ID protection objectives and loss functions. Sec. \ref{sec:adaptive_attacks} introduces the data augmentation approach to enhance the robustness.

\subsection{Overall Method}
\label{sec:overall_method}

Fig.~\ref{fig:architecture} details the pipeline of our method. For an arbitrary-sized input image $x$ containing a face, it is resized to 224$\times$224 and sent into the adversarial noise encoder model, whose output is an additive adversarial noise that can be resized back and perturbed onto the original image for protection. During adversarial noise encoder training, this protected image is further passed through face feature extraction pipelines to obtain the loss functions that aim at creating maximum differences in the feature space.

\subsection{Noise Encoder}
\label{sec:adv_noise_encoder}

The network takes a 224$\times$224 RGB image as input and outputs three-channel adversarial perturbations. The perturbations are projected to a range of $[-1, 1]$, then denormalized $[-\epsilon, \epsilon]$, resized to the original image dimensions, and added to the image. We utilize a Vision Transformer (ViT) to generate the adversarial noise, with both input and output dimensions set to 224$\times$224. Empirically, we find that adding an additional prior mask channel, indicating the location of the face, aids in training. This face localization mask is generated by the InsightFace pipeline based on facial landmark points, which specify the region of the face within the image. The mask is concatenated to the image as the fourth input channel, eliminating the need for the network to learn face localization ability from scratch, thus reducing training difficulty.

\subsection{Adversarial ID Protection}
\label{sec:IP_protection_obj}
Since generative networks' only knowledge of faces comes from feature embeddings, to achieve ID protection, we can make the extracted features from protected images significantly different from the originals, thereby preventing generation models from accessing correct facial features. Based on this principle, we analyze the specific pipelines of InstantID, IP-Adapter, IP-Adapter-Plus, and PhotoMaker to design targeted attack objectives and loss functions.

\textbf{InstantID} obtains facial features in two steps, as shown in Fig. \ref{fig:architecture}. It first aligns the face to a pre-defined location, before feeding the aligned face into an ArcFace feature extractor to obtain facial information. We choose to fool the ArcFace model, minimizing the cosine similarity between the ArcFace features of the original \textit{v.s.} protected images.

\textbf{IP-Adapter}, \textbf{IP-Adapter-Plus}, and \textbf{PhotoMaker} all rely on facial features extracted by the CLIP vision encoder, albeit using different versions of CLIP vision. IP-Adapter and PhotoMaker utilize CLIP Vision's output, while IP-Adapter-Plus takes features before CLIP Vision's second-to-last layer. As illustrated in Fig. \ref{fig:architecture}, embeddings from different layers offer various attack surfaces. We select victim embeddings for attack according to three principles: (1) blocking all potential pathways, ensuring that any information flow from left to right passes through at least one victim embedding, ensuring all information flow would be disrupted; (2) choosing features as early as possible in the network (towards the left in Fig. \ref{fig:architecture}), to reduce the backpropagation path length and thus simplify optimization; and (3) targeting embeddings with dense semantic information for more effective manipulation.

An example of an unsuitable target is the last hidden layer feature of CLIP Vision. While modifying this feature could influence all three models, its huge dimension (257$\times$1280) and lack of semantic density make it highly challenging for optimization to converge. Ultimately, we selected the embeddings marked red in Fig. \ref{fig:architecture} as the primary attack targets. The objective, similar to that of InstantID, is to maximize the cosine similarity between the perturbed embedding and the original embedding, effectively aligning them post-attack. Therefore, the final adversarial loss is a weighted average of all losses, given by
\begin{equation}
    L_\text{adv} = \sum_{\text{Custom Model }i} \alpha_i \cdot \text{cossim}(e_i, e_i'),
\end{equation}
where $e_i$ and $e_i'$ denote the face embeddings of the clean and protected images, respectively.

\subsection{Imperceptibility}
\label{sec:loss_functions}
To minimize the visual impact on the image quality, we enforce an $\ell_1$ regularization on the predicted adversarial noise $\delta$. 
We further introduce an auxiliary penalty on any noise values exceeding the $\epsilon$-ball boundary.
Together, these terms constitute our regularization loss, given by:
\begin{equation}
    L_\text{reg} = \beta_1 \cdot \lVert\delta\rVert_1 + \beta_2 \cdot \lVert\delta-\text{clip}_{\pm \epsilon}(\delta) \rVert_1.
\end{equation}
The final loss is thus the sum of adversarial loss and regularization, given by:
\begin{equation}
    L = L_\text{adv} + L_\text{reg}.
    \label{eq:overall_loss}
\end{equation}

\subsection{Robustness}
\label{sec:adaptive_attacks}
As shown in Fig. \ref{fig:architecture}, the InstantID pipeline involves a step of facial alignment aiming to align any faces to a predefined standard face for better feature extraction.
The alignment process includes detecting the face and key points $\Tilde{p}$, calculating the affine matrix $A$ with standard face key points $\Tilde{p}'$, and aligning them. This process is computationally expensive.
To reduce computational cost and simplify training, we pre-compute $A$ for each image and keep it fixed throughout training. 
However, during inference, perturbations carried by protected images may slightly alter the pre-computed affine matrix $A$, leading to minor misalignment that weakens the effectiveness the adversarial perturbations. 

To adapt to this misalignment as well as being computationally efficient, we leverage the existing affine transformation operation and introduce small gaussian noise to $A$ during training to simulate face transformations. The noise level is empirically set to $\sigma=0.003$. The face alignment transformation is therefore given by
\begin{equation}
    \Tilde{p}' = \left( A + N(0, \sigma I)\right) \Tilde{p}.
\end{equation}
We incorporate this augmentation only in the InstantID branch. Our experiments reveal that it also enhances the robustness of other branches, being resilient to various common image transformations.


\section{Experiments}
\label{sec:exper}

\subsection{Experimental Setup}

\textbf{Training Data.} For practicality, we selected 190,000 uncropped images from the CelebA dataset~\cite{celeba} as the training set. These images have arbitrary sizes and aspect ratios. The test set includes two subsets: 50 unseen images from CelebA and 50 images from VGG Face \cite{vggface}. To ensure compatibility with baseline methods, these images were resized to 512$\times$512 and 224$\times$224, respectively.

\textbf{Training Details.} We trained a ViT-S/8 adversarial noise encoder from scratch using Eq. \ref{eq:overall_loss} as the loss function. We employed a three-stage curriculum learning approach, progressively shrinking the $\epsilon$-ball and balancing the different losses by adjusting the hyperparameters (presented in Table \ref{tab:training_stages}). Training began with a 2,500-step linear warm-up, gradually increasing the learning rate from zero. Gradient clipping was applied during training to cap the $\ell_2$ norm of gradients at a maximum of 10. The model was trained with a batch size of 112 on four 80GB NVIDIA H100 GPUs for 10 days.

\begin{table}[]
\centering
\caption{Curriculum learning stages with corresponding hyperparameters. $\eta$ represents learning rate. $\alpha_1$ to $\alpha_4$ are weights for IP-Adapter, IP-Adapter-Plus, PhotoMaker, InstantID, respectively.}
\label{tab:training_stages}
\setlength{\tabcolsep}{4pt}
\resizebox{\linewidth}{!}{%
\begin{tabular}{cccccccccc}
\toprule
\textbf{Stage} & \textbf{Epochs} & \bm{$\alpha_1$} & \bm{$\alpha_2$} & \bm{$\alpha_3$} & \bm{$\alpha_4$} & \bm{$\beta_1$} & \bm{$\beta_2$} & \bm{$\epsilon$} & \bm{$\eta$} \\ \hline & \\[-2.1ex]
1              & 120             & 2/37                & 14/37               & 20/37               & 1/37                & 9e-3               & 1e-3               & 0.05                & 1e-2            \\
2              & 20              & 4/16                & 2/16                & 9/16                & 2/16                & 1.8e-3             & 2e-4               & 0.04                & 2e-5            \\
3              & 20              & 2/14                & 2/14                & 9/14                & 1/14                & 4.5e-4             & 5e-5               & 0.035               & 2e-5            \\ \bottomrule
\end{tabular}%
}
\end{table}

\textbf{Evaluation process and metrics.} To evaluate the protection performance of different methods, for each face-containing image in the dataset (denoted as NP ref image), we apply different protection methods (including our IDProtector) to generate its protected counterpart (denoted as P ref image). Throughout all experiments, we constrain the maximum pixel perturbation to 9/255 (an $\epsilon$-ball of 0.035), following common practice in adversarial machine learning \cite{saberi2023robustness}. Subsequently, both P and NP ref images are fed into customization generation models (such as InstantID) to obtain generated images (denoted as P/NP gen images).

We evaluate the identify shift caused by different protection methods through \textbf{Identity Score Matching (ISM)}, which is calculated as the cosine similarity between facial features of NP images and generated images extracted by ArcFace \cite{arcface}. Lower ISM values indicate better protection as they suggest a greater identity shift between original (NP ref) and generated (P gen) faces.

Since protected images may result in no detectable faces being generated, we also report the \textbf{Face Detection Rate (FDR)} using InsightFace detector, as ISM can only be computed when faces are detected in both images. Additionally, to ensure comprehensive evaluation, we measure the perceptual quality of each generated face using SER-FIQ \cite{terhorst2020ser}.

\begin{figure*}
    \centering
    \includegraphics[width=\linewidth]{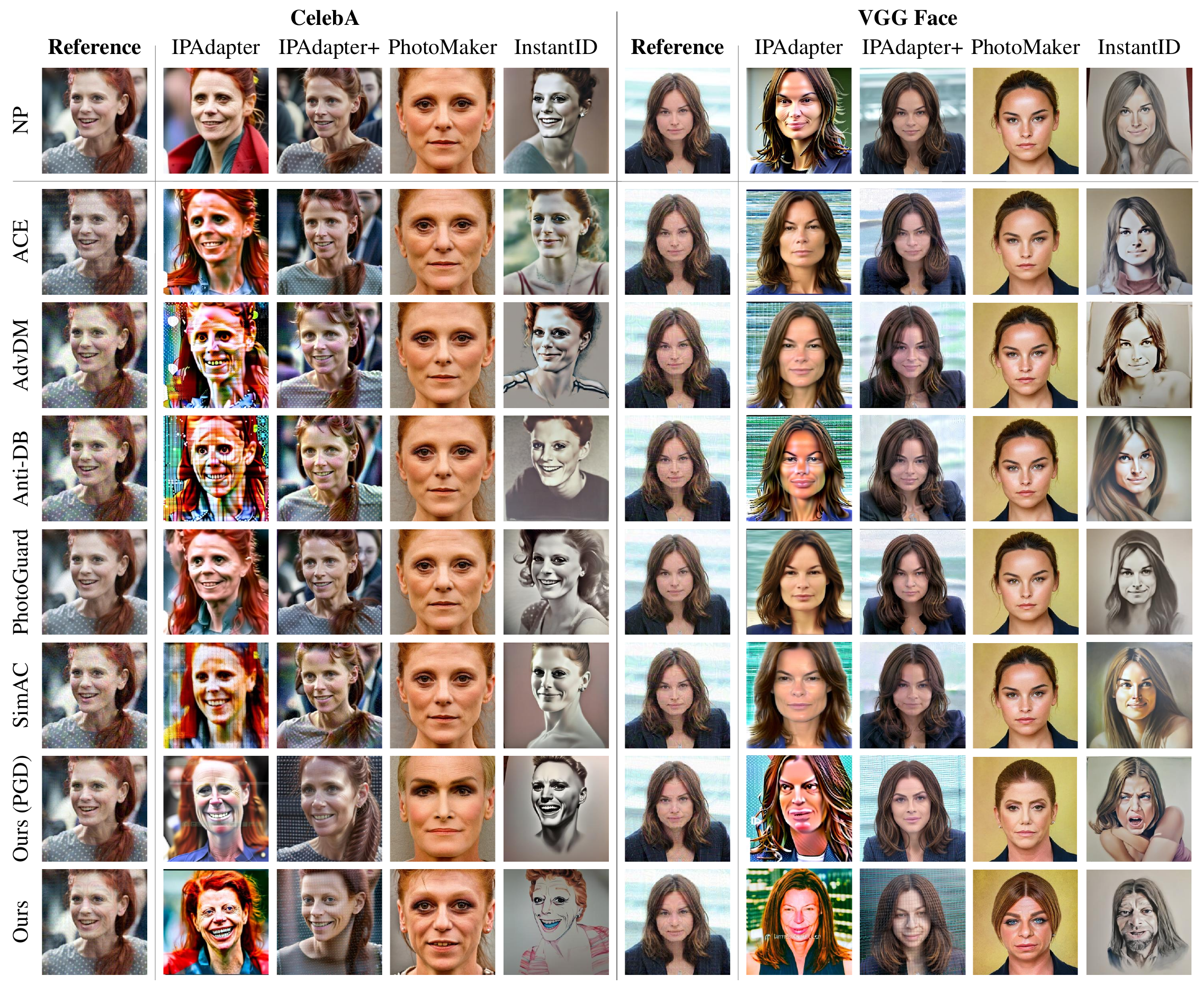}
    \caption{Qualitative comparison with baseline. Images protected by our method cannot be used to generate similar faces. Compared to baseline protection methods, under the same perturbation bound of 9/255 ($\epsilon=3.5\%$), ours causes more significant facial differences.}
    \label{fig:qualitative_with_baseline}
\end{figure*}

\begin{table*}[]
\caption{Comparison of identity protection performance against baseline methods. Protection performance is evaluated against IP-Adapter, IP-Adapter-Plus, PhotoMaker, and InstantID. ``No Protect" denotes direct use of the original (NP ref) images for identity-preserving generation. Experiments were conducted on five prompts per dataset, and average results are reported.}
\label{tab:main_quantitative}
\setlength{\tabcolsep}{4pt}
\resizebox{\textwidth}{!}{%
\begin{tabular}{llccccccccccccccc}
\toprule
\multicolumn{1}{c}{\multirow{2}{*}{\textbf{Dataset}}}                         & \multicolumn{1}{c}{\multirow{2}{*}{\textbf{\begin{tabular}[c]{@{}c@{}}Protection\\ Method\end{tabular}}}} & \multicolumn{3}{c}{\textbf{IP-Adapter}}                                               & \textbf{} & \multicolumn{3}{c}{\textbf{IP-Adapter-Plus}}                                          & \textbf{} & \multicolumn{3}{c}{\textbf{PhotoMaker}}                                               & \textbf{} & \multicolumn{3}{c}{\textbf{InstantID}}                                                \\ \cline{3-5} \cline{7-9} \cline{11-13} \cline{15-17} 
\multicolumn{1}{c}{}                                                          & \multicolumn{1}{c}{}                                                                                      & \textbf{FDR $\downarrow$} & \textbf{ISM $\downarrow$} & \textbf{SER-FIQ $\downarrow$} & \textbf{} & \textbf{FDR $\downarrow$} & \textbf{ISM $\downarrow$} & \textbf{SER-FIQ $\downarrow$} & \textbf{} & \textbf{FDR $\downarrow$} & \textbf{ISM $\downarrow$} & \textbf{SER-FIQ $\downarrow$} & \textbf{} & \textbf{FDR $\downarrow$} & \textbf{ISM $\downarrow$} & \textbf{SER-FIQ $\downarrow$} \\ \hline & \\[-2.1ex]
\multirow{8}{*}{\textbf{CelebA}}                                              & \textbf{No Protect}                                                                                       & 1.000                     & 0.176                     & 0.641                         &           & 1.000                     & 0.240                     & 0.663                         &           & 0.664                     & 0.265                     & 0.605                         &           & 0.996                     & 0.775                     & 0.774                         \\
                                                                              & \textbf{ACE}                                                                                              & 1.000                     & 0.159                     & 0.640                         &           & 1.000                     & 0.231                     & 0.681                         &           & 0.636                     & 0.268                     & 0.594                         &           & 1.000                     & 0.745                     & 0.758                         \\
                                                                              & \textbf{AdvDM}                                                                                            & 0.988                     & 0.157                     & 0.630                         &           & 1.000                     & 0.226                     & 0.666                         &           & 0.640                     & 0.255                     & \textbf{0.590}                &           & 1.000                     & 0.757                     & 0.757                         \\
                                                                              & \textbf{Anti-DB}                                                                                          & 1.000                     & 0.155                     & 0.615                         &           & 1.000                     & 0.222                     & 0.676                         &           & 0.640                     & 0.255                     & 0.599                         &           & 1.000                     & 0.753                     & 0.752                         \\
                                                                              & \textbf{PhotoGuard}                                                                                       & 1.000                     & 0.152                     & 0.621                         &           & 1.000                     & 0.223                     & 0.670                         &           & 0.652                     & 0.259                     & 0.592                         &           & 1.000                     & 0.758                     & 0.767                         \\
                                                                              & \textbf{SimAC}                                                                                            & 1.000                     & 0.162                     & 0.627                         &           & 1.000                     & 0.217                     & 0.683                         &           & 0.652                     & 0.253                     & 0.599                         &           & 1.000                     & 0.752                     & 0.755                         \\
                                                                              & \textbf{Ours (PGD)}                                                                                         & \textbf{0.844}            & \textbf{0.024}            & \textbf{0.544}                &           & \textbf{0.912}            & \textbf{0.072}            & 0.718                         &           & \textbf{0.112}            & \textbf{0.094}            & 0.639                         &           & \textbf{0.996}            & \textbf{-0.166}           & 0.721                         \\
                                                                              & \textbf{Ours}                                                                                             & 0.952                     & 0.060                     & 0.688                         &           & 0.996                     & 0.114                     & \textbf{0.644}                &           & 0.672                     & 0.123                     & 0.622                         &           & 1.000                     & 0.231                     & \textbf{0.613}                \\ \hline & \\[-2.1ex]
\multirow{8}{*}{\textbf{\begin{tabular}[c]{@{}l@{}}VGG \\ Face\end{tabular}}} & \textbf{No Protect}                                                                                       & 1.000                     & 0.201                     & 0.672                         &           & 1.000                     & 0.236                     & 0.698                         &           & 0.632                     & 0.254                     & 0.600                         &           & 0.996                     & 0.764                     & 0.828                         \\
                                                                              & \textbf{ACE}                                                                                              & 0.988                     & 0.154                     & 0.617                         &           & 1.000                     & 0.209                     & 0.683                         &           & 1.000                     & 0.209                     & 0.683                         &           & 1.000                     & 0.703                     & 0.829                         \\
                                                                              & \textbf{AdvDM}                                                                                            & 1.000                     & 0.147                     & \textbf{0.597}                &           & 1.000                     & 0.215                     & 0.693                         &           & 1.000                     & 0.215                     & 0.693                         &           & 1.000                     & 0.706                     & 0.820                         \\
                                                                              & \textbf{Anti-DB}                                                                                          & 1.000                     & 0.138                     & \textbf{0.597}                &           & 1.000                     & 0.210                     & 0.678                         &           & \textbf{0.624}            & 0.238                     & \textbf{0.599}                &           & 1.000                     & 0.707                     & 0.811                         \\
                                                                              & \textbf{PhotoGuard}                                                                                       & 1.000                     & 0.159                     & 0.643                         &           & 1.000                     & 0.217                     & 0.686                         &           & \textbf{0.624}            & 0.229                     & 0.612                         &           & \textbf{0.996}            & 0.691                     & 0.799                         \\
                                                                              & \textbf{SimAC}                                                                                            & 1.000                     & 0.150                     & 0.605                         &           & 1.000                     & 0.202                     & \textbf{0.677}                &           & \textbf{0.624}            & 0.228                     & 0.644                         &           & 1.000                     & 0.707                     & 0.826                         \\
                                                                              & \textbf{Ours (PGD)}                                                                                         & \textbf{0.968}            & \textbf{0.043}            & 0.609                         &           & 1.000                     & 0.202                     & 0.740                         &           & \textbf{0.624}            & 0.228                     & 0.630                         &           & 1.000                     & 0.707                     & 0.769                         \\
                                                                              & \textbf{Ours}                                                                                             & 1.000                     & 0.105                     & 0.737                         &           & 1.000                     & \textbf{0.138}            & 0.687                         &           & 0.632                     & \textbf{0.126}            & 0.614                         &           & 1.000                     & \textbf{0.232}            & \textbf{0.713}                \\ \bottomrule
\end{tabular}%
}
\end{table*}

\subsection{Comparison with Baselines}

\textbf{Protection Performance.} We compare our method with the following baseline methods: ACE \cite{ace}, AdvDM \cite{advdm}, Anti-DreamBooth \cite{antidreambooth}, PhotoGuard \cite{photoguard}, and SimAC \cite{simac}. Most baselines aim to reduce the quality of generated images or hinder the diffusion denoising process, but we aim to reduce the similarity of generated identities so as to directly disrupt identity-preserving generation. To compare our method with baselines, we first applied each protection method to generate protected versions of test images. We then used both the original images (NP) and the protected versions as references in customized generation tasks with four models: IP-Adapter, IP-Adapter-Plus, PhotoMaker, and InstantID. Tab. \ref{tab:main_quantitative} shows that, for the first three methods, while baseline methods reduced ISM by an average of 0.02, with a maximum reduction of 0.06, our method achieved an ISM reduction between 0.1 and 0.14, demonstrating a larger identity shift. On InstantID, our IDProtector demonstrated even more pronounced protection: when baseline methods reduced facial similarity by less than 0.1, our method reduced similarity by over 0.4 across both datasets. Additionally, FDR results validate the reliability of ISM as FDR reflects the percentage of images with a face detected and contributes to ISM calculations. Overall, these results underscore the superior identity protection performance of our method compared to existing approaches. Together, these highlight the superior identity protection performance of our approach.

Following previous study \cite{antidreambooth}, we evaluate SER-FIQ as a quality metric of generated (P/NP gen) images. Although our method has a noticeable impact on image generation quality (as shown in Fig. \ref{fig:qualitative_with_baseline}), we observed inconsistencies between the SER-FIQ scores and qualitative assessments. This discrepancy suggests that a more robust and reliable quality metric may be required to accurately measure the quality of generated images in future research.

We also adapted our method to a per-image tuning variant, employing PGD \cite{pgd} as the optimization technique. We observed that PGD achieves a more substantial ISM reduction across most methods, which aligns with our expectations since per-image tuning can better leverage the specific characteristics of each individual image. However, a notable drawback of PGD is the prolonged optimization time, which we will address in the following section.

The protection effects are visualized in Fig. \ref{fig:qualitative_with_baseline}. The baseline methods can protect against some of the generation methods. For example, AdvDM and Anti-DreamBooth cause IP-Adapter to produce faces with color block artifacts, while several outputs from IP-Adapter-Plus exhibit unnatural patterns and shadows. However, the baselines are ineffective in protecting against PhotoMaker and InstantID. With IDProtector's protection, faces generated by all four methods appear significantly altered. In some cases, even the gender of the person is unidentifiable. This fulfills our protection goal: protected images could not be used for identity-preserving generation.

\textbf{Time and Image Quality.} 
Our IDProtector is fast to protect an image with less degradation to image quality. As compared in Table \ref{tab:attack_metric}, the PSNR and SSIM of our method are higher than the baselines, indicating less visual impact. The baseline methods require an average of 23.3 to 323 seconds per image for protection. Using PGD with our optimization objective would take over an hour on average. In contrast, our ViT-based IDProtector achieves an average protection time of only 0.173 seconds per image -- less than 1\% of the fastest baseline. Note that the forward-feeding ViT supports better parallelization compared with PGD, the time cost could further decrease with mini-batch inference.

\begin{table}[]
\caption{Comparison of time cost and quality impact with an epsilon-budget of 9/255 ($\pm$3.5\%). PSNR and SSIM are calculated between clean (NP ref) and perturbed (P ref) image pairs.}
\label{tab:attack_metric}
\setlength{\tabcolsep}{4pt}
\resizebox{\linewidth}{!}{%
\begin{tabular}{lccccccc}
\toprule
                   & \textbf{ACE} & \textbf{AdvDM} & \textbf{\begin{tabular}[c]{@{}c@{}}Anti-\\ DB\end{tabular}} & \textbf{\begin{tabular}[c]{@{}c@{}}Photo-\\ Guard\end{tabular}} & \textbf{SimAC} & \textbf{\begin{tabular}[c]{@{}c@{}}Ours\\ PGD\end{tabular}} & \textbf{Ours}  \\ \hline & \\[-2.1ex]
\textbf{Time (s)}  & 87.0         & 51.9           & 261                                                         & 23.3                                                            & 323            & 4089                                                        & \textbf{0.173} \\
\textbf{PSNR (dB)} & 31.06        & 31.70          & 31.70                                                       & 31.10                                                           & 32.37          & \textbf{32.94}                                              & 32.15          \\
\textbf{SSIM (dB)} & 0.741        & 0.766          & 0.765                                                       & 0.757                                                           & 0.791          & \textbf{0.845}                                              & 0.809          \\ \bottomrule
\end{tabular}%
}
\end{table}

\begin{table}[]
\caption{Generalization of ID protection by our IDProtector on unseen generators. IPA-FID/+/P/XL denotes IP-Adapter-FaceID/Plus/Portrait/SDXL, with NP and P indicating no and with protection, respectively.}
\label{tab:generalisation}
\setlength{\tabcolsep}{3pt}
\resizebox{\linewidth}{!}{%
\begin{tabular}{llcccccccc}
\toprule
                                                                                          &              & \textbf{\begin{tabular}[c]{@{}c@{}}IPA-\\ FID\end{tabular}} & \textbf{\begin{tabular}[c]{@{}c@{}}IPA-\\ FID+\end{tabular}} & \textbf{\begin{tabular}[c]{@{}c@{}}IPA-\\ FIDP\end{tabular}} & \textbf{\begin{tabular}[c]{@{}c@{}}IPA-\\ FIDXL\end{tabular}} & \textbf{\begin{tabular}[c]{@{}c@{}}Flux-\\ IPA\end{tabular}} & \textbf{\begin{tabular}[c]{@{}c@{}}Story-\\ Maker\end{tabular}} & \textbf{\begin{tabular}[c]{@{}c@{}}Mid-\\ journey\end{tabular}} & \textbf{\begin{tabular}[c]{@{}c@{}}Jing\\ Gou\end{tabular}} \\ \hline & \\[-2.1ex]
\multirow{2}{*}{\textbf{ISM $\downarrow$}}                                                & \textbf{NP}  & 0.565                                                       & 0.644                                                        & 0.129                                                        & 0.523                                                         & 0.054                                                        & 0.572                                                           & 0.199                                                           & 0.672                                                       \\
                                                                                          & \textbf{P}   & 0.153                                                       & 0.217                                                        & 0.052                                                        & 0.132                                                         & 0.029                                                        & 0.193                                                           & 0.098                                                           & 0.131                                                       \\
\multirow{2}{*}{\textbf{\begin{tabular}[c]{@{}l@{}}SER-\\ FIQ $\downarrow$\end{tabular}}} & \textbf{NP}  & 0.706                                                       & 0.751                                                        & 0.720                                                        & 0.402                                                         & 0.728                                                        & 0.699                                                           & 0.568                                                           & 0.791                                                       \\
                                                                                          & \textbf{P}   & 0.803                                                       & 0.730                                                        & 0.658                                                        & 0.667                                                         & 0.673                                                        & 0.734                                                           & 0.627                                                           & 0.805                                                       \\ \bottomrule
\end{tabular}%
}
\end{table}

\begin{table*}[]
\caption{Robustness evaluation of IDProtector under image distortions. NP denotes non-protected images for identity-preserving generation, while P represents protected but non-distorted images, providing an ablation. Distortion averages exclude NP and P.}
\label{tab:robustness}
\setlength{\tabcolsep}{3pt}
\resizebox{\textwidth}{!}{%
\begin{tabular}{lcclcclcclcclccccccccccc}
\toprule
                  & \multicolumn{11}{c}{\textbf{IDProtector (ours, ViT)}}                                                                                                                                                                                  &  & \multicolumn{11}{c}{\textbf{Adaptive PGD}}                                                                                                                                                                                                                        \\ \cline{2-12} \cline{14-24} 
                  & \multicolumn{2}{c}{\textbf{IP-Adapter}}               &  & \multicolumn{2}{c}{\textbf{IP-Adapter+}}              &  & \multicolumn{2}{c}{\textbf{PhotoMaker}}               &  & \multicolumn{2}{c}{\textbf{InstantID}}                &  & \multicolumn{2}{c}{\textbf{IP-Adapter}}               & \textbf{} & \multicolumn{2}{c}{\textbf{IP-Adapter+}}              & \textbf{} & \multicolumn{2}{c}{\textbf{PhotoMaker}}               & \textbf{} & \multicolumn{2}{c}{\textbf{InstantID}}                \\
                  & \textbf{FDR $\downarrow$} & \textbf{ISM $\downarrow$} &  & \textbf{FDR $\downarrow$} & \textbf{ISM $\downarrow$} &  & \textbf{FDR $\downarrow$} & \textbf{ISM $\downarrow$} &  & \textbf{FDR $\downarrow$} & \textbf{ISM $\downarrow$} &  & \textbf{FDR $\downarrow$} & \textbf{ISM $\downarrow$} &           & \textbf{FDR $\downarrow$} & \textbf{ISM $\downarrow$} &           & \textbf{FDR $\downarrow$} & \textbf{ISM $\downarrow$} &           & \textbf{FDR $\downarrow$} & \textbf{ISM $\downarrow$} \\ \hline
\textbf{NP}       & 1.000                     & 0.176                     &  & 1.000                     & 0.240                     &  & 0.664                     & 0.265                     &  & \textbf{0.996}            & 0.775                     &  & 0.844                     & 0.024                     &           & \textbf{0.912}            & 0.072                     &           & 0.112                     & 0.094                     &           & \textbf{0.996}            & -0.166                    \\ \hline
\textbf{P}        & \textbf{0.952}            & \textbf{0.060}            &  & 0.996                     & 0.114                     &  & 0.672                     & \textbf{0.123}            &  & 1.000                     & 0.231                     &  & 0.844                     & 0.024                     &           & \textbf{0.912}            & 0.072                     &           & 0.112                     & 0.094                     &           & \textbf{0.996}            & -0.166                    \\
\textbf{P+Affine} & 1.000                     & 0.068                     &  & 1.000                     & 0.106                     &  & 0.656                     & 0.135                     &  & 1.000                     & \textbf{0.216}            &  & 0.844                     & 0.064                     &           & 0.976                     & 0.081                     &           & 0.112                     & -0.009                    &           & 1.000                     & -0.147                    \\
\textbf{P+JPEG}   & 0.964                     & 0.069                     &  & 1.000                     & \textbf{0.097}            &  & 0.660                     & 0.138                     &  & 1.000                     & 0.241                     &  & 0.784                     & \textbf{0.015}            &           & 0.948                     & \textbf{0.062}            &           & \textbf{0.108}            & \textbf{-0.012}           &           & 1.000                     & -0.161                    \\
\textbf{P+Crop}   & 1.000                     & 0.072                     &  & 1.000                     & 0.108                     &  & \textbf{0.652}            & 0.132                     &  & 1.000                     & 0.216                     &  & \textbf{0.752}            & 0.058                     &           & 0.940                     & 0.070                     &           & 0.108                     & 0.020                     &           & 1.000                     & \textbf{-0.167}           \\
\textbf{P+Noisy}  & \textbf{0.952}            & 0.069                     &  & \textbf{0.984}            & 0.114                     &  & 0.656                     & 0.132                     &  & 1.000                     & 0.252                     &  & 0.832                     & 0.023                     &           & 0.932                     & 0.072                     &           & 0.112                     & -0.010                    &           & 1.000                     & -0.130                    \\
\textbf{Avg (P+)} & 0.979                     & 0.070                     &  & 0.996                     & 0.106                     &  & 0.656                     & 0.134                     &  & 1.000                     & 0.231                     &  & 0.803                     & 0.040                     &           & 0.949                     & 0.071                     &           & 0.110                     & -0.003                    &           & 1.000                     & -0.151                    \\ \bottomrule
\end{tabular}%
}
\end{table*}

\begin{figure}
    \centering
    \includegraphics[width=\linewidth]{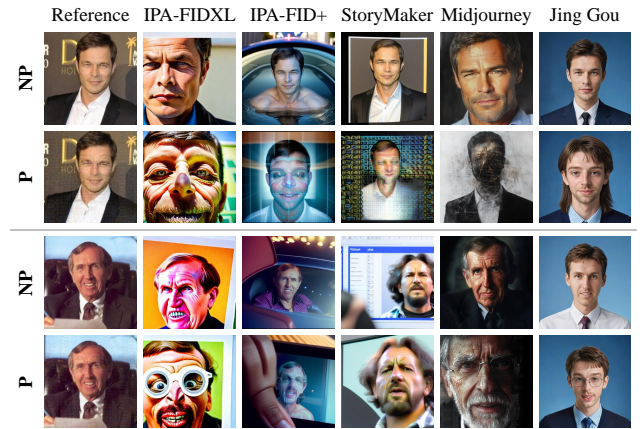}
    \caption{ID protection performance on unseen generators.}
    \label{fig:generalisation}
\end{figure}

\begin{figure}
    \centering
    \includegraphics[width=\linewidth]{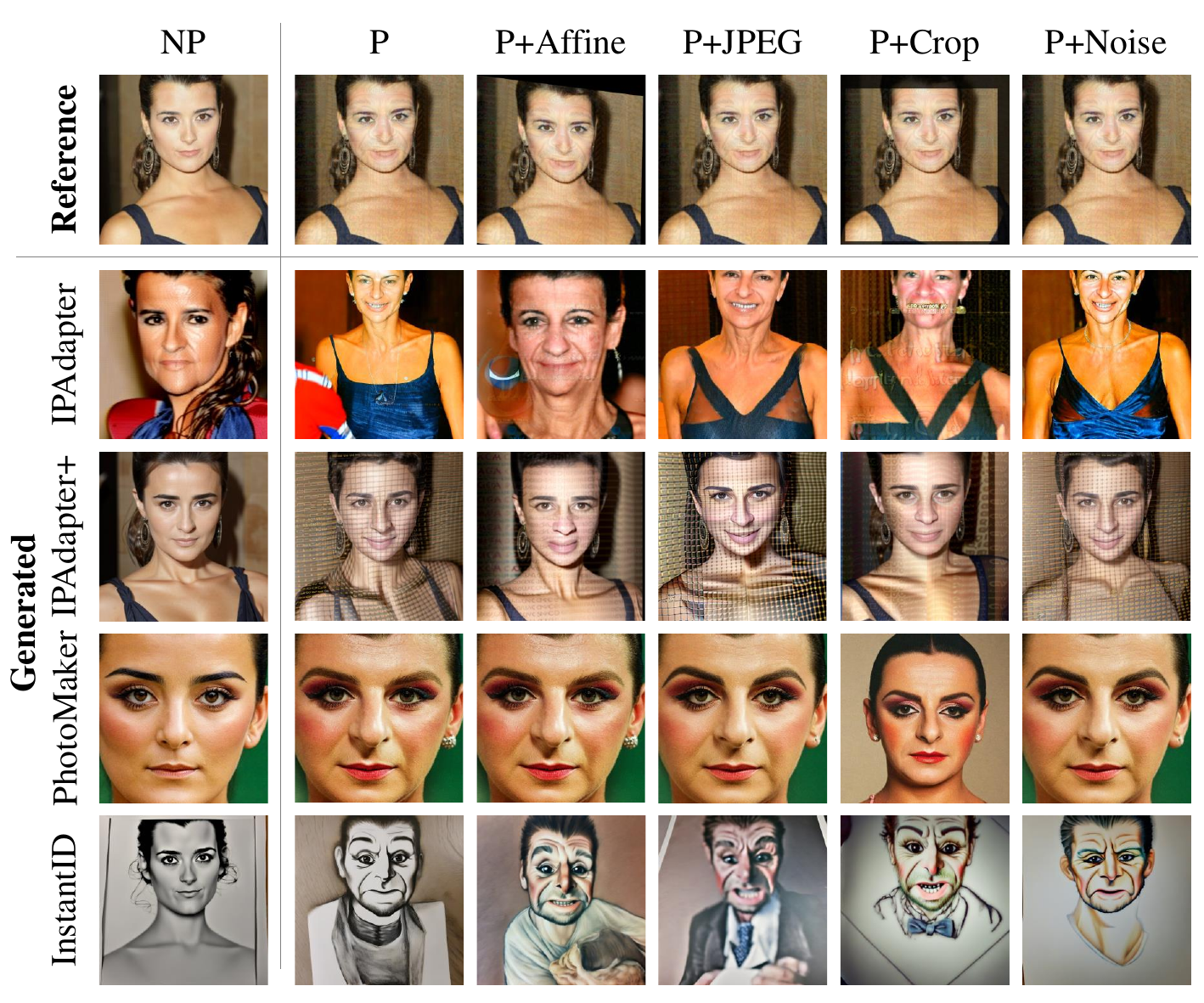}
    \caption{Robustness evaluation of IDProtector under various distortions. Despite geometric distortions like affine transformations, IDProtector effectively misleads all four models into generating dissimilar faces or even artifacts.}
    \label{fig:robustness}
\end{figure}

\subsection{Generalization to Unseen Data and Models}

\textbf{Generalization to unseen datasets.} We evaluated our method on the VGG Face dataset that was unseen during training. All following experiments are conducted on the CelebA test set unless otherwise specified. As shown in Tab. \ref{tab:main_quantitative}, the ISM reduction achieved by our method shows minimal variation across datasets. For example, our IDProtector reduces the ISM of IP-Adapter by approximately 0.1 and InstantID by around 0.55 on both datasets. This result suggests that our method generalises effectively beyond the CelebA dataset, demonstrating protection capabilities that are not restricted to in-domain faces.

\textbf{Generalization to unseen customization models.} Beyond the four customization models seen during training, we tested protected images on a wider range of customization models. Across these models, our protection approach achieves an average ISM reduction of 0.31, with a maximum reduction of 0.54. Notably, our method remains effective against proprietary commercial models such as Midjourney (0.10 reduction) and Jing Gou (0.54 reduction), underscoreing its ability to generalise to unseen customization models. 

These protection effects are visualised in Fig. \ref{fig:generalisation}. As shown, protected images used in these generators cause noticeable quality degradation, including altered facial shapes (e.g., IP-Adapter-FaceID-XL), reduced prompt-following ability, and image contamination (e.g., Midjourney). The common result is a decreased resemblance between the generated face and the reference identity.

\subsection{Robustness Evaluation}

Our method demonstrates robustness against various distortions. For protected (P) images, we applied several distortions, including JPEG compression (quality=85), random cropping that removes up to 20\% of the region, noise perturbation with a variance of 100, and an affine transformation. The affine transformation was implemented as $\Tilde{q}=(A+\mathcal{N}(0, 0.05^2))\Tilde{p}$, where the matrix $A$ is an identity matrix. As shown in Tab. \ref{tab:robustness}, the ISM values under these distortions remain comparable to those obtained without distortion, indicating that our protection remains effective under distortion. Fig. \ref{fig:robustness} provides a visual comparison, where even under various distortions, InstantID fails to reconstruct the correct female face, generating a male portrait instead. This demonstrates that the common distortions do not compromise the efficacy of our protection method.



 
\section{Conclusion and Limitation}


In this paper, we introduce IDProtector, a method designed to protect facial identities from encoder-based identity-preserving generation. By training a noise encoder to add adversarial perturbations onto images, IDProtector can protect an image within 0.2s. Our method achieves significantly lower face similarity in generated images than existing protection methods. 
IDProtector also demonstrates strong practical applicability.
Its protection remains effective under distortions such as affine transformations and JPEG compression. Furthermore, IDProtector not only shows effectiveness against closed-source models, such as Midjourney, but also unifies protection against multiple ID-preserving models, setting a new standard for privacy protection against generative AI.

A limitation of our method is that the adversarial noise is not fully invisible, necessitating a trade-off between attack effectiveness and Imperceptibility. Future efforts will focus on enhancing the Imperceptibility of the noise while maintaining its robustness.


\clearpage
{
    \small
    \bibliographystyle{ieeenat_fullname}
    \bibliography{main}
}

\setcounter{page}{1}
\maketitlesupplementary
\renewcommand\thesection{\Alph{section}}
\setcounter{section}{0}

\section{Details of Prior Channels}

As introduced in Sec. \ref{sec:adv_noise_encoder}, IDProtector incorporates additional prior channels alongside the RGB channels to reduce the network's learning burden. Here, we elaborate on the details of these two prior channels and explain how they facilitate the network's learning process.

\subsection{Face Localization Prior}

\begin{figure}
    \centering
    \includegraphics[width=\linewidth]{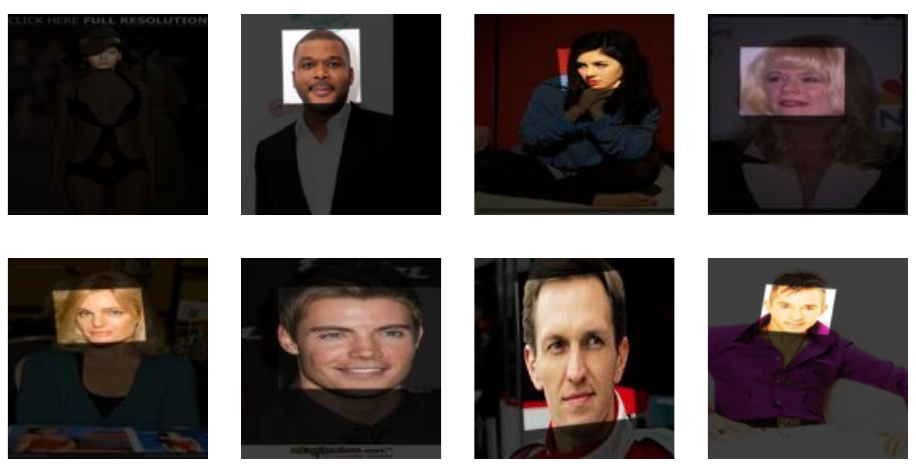}
    \caption{Visualization of face localization prior. The prior channel is overlayed as the difference in brightness levels onto the original RGB images. Bright regions indicate 1s in the face localization prior channel, which would be pixels visible to ArcFace. Dark regions indicate 0s in the prior channel, which would be discarded before reaching ArcFace.}
    \label{fig:vis_face_prior}
\end{figure}

The first prior channel is a face localization prior, implemented as a binary mask channel with values of 0 or 1, which indicates the facial region within the image (bright regions as visualized in Fig. \ref{fig:vis_face_prior}). This prior channel explicitly marks the regions that InstantID's preprocessing step will preserve, thereby eliminating the need for the network to learn face localization capabilities independently, thus reducing the training burden.

\begin{figure}
    \centering
    \includegraphics[width=\linewidth]{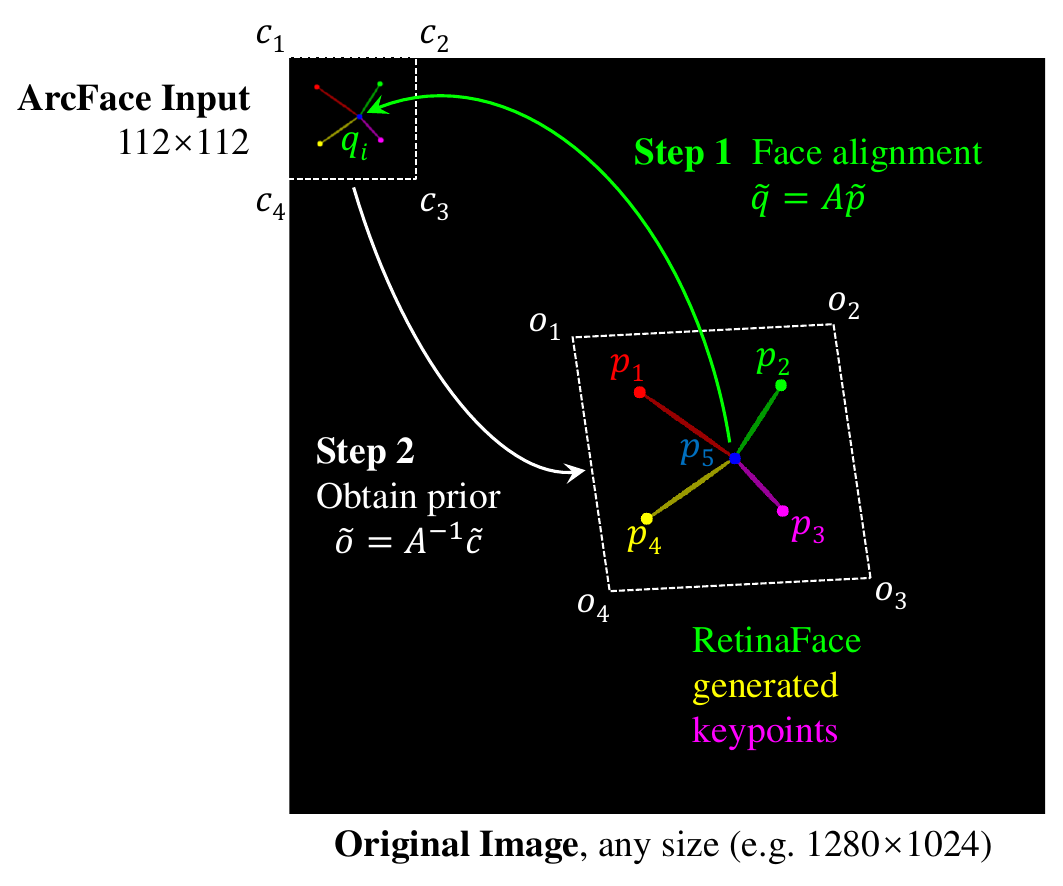}
    \caption{Visualization of ArcFace alignment and how the face localization prior is obtained.}
    \label{fig:affine_visualisation}
\end{figure}

The face localization region is defined by a quadrilateral whose vertices are computed through the following process. Initially, the three-channel 224×224 RGB image is processed by RetinaFace, which outputs five facial landmark points, denoted as $\Tilde{\textbf{p}}_1, ..., \Tilde{\textbf{p}}_5$. These landmarks correspond to the eyes, nose tip, and mouth corners. These points are used to align the facial features with ArcFace's required input format, which defines five predefined landmark locations $\Tilde{\textbf{q}}_1, ..., \Tilde{\textbf{q}}_5$. The alignment is achieved through an affine transformation, as illustrated in Figure X, resulting in a 112×112 aligned facial image. The affine transformation is given by
\begin{equation}
    \Tilde{\textbf{q}} = A \Tilde{\textbf{p}},
\end{equation}
where $A$ represents the affine transformation matrix. This matrix is computed by minimizing the transformation fitting error:
\begin{equation}
    A = \arg \min_A \sum_{\text{landmark} i} \lVert \Tilde{\textbf{q}}_i - A\Tilde{\textbf{p}}_i\rVert_2^2,
\end{equation}
which is solved using least squares fitting. Subsequently, the 112×112 region in the upper-left corner is cropped and fed into ArcFace as the aligned face. To mark this cropped region in the original image (which constitutes our face localization prior), we perform an inverse affine transformation. Denoting the coordinates of the square region's vertices as $\Tilde{\textbf{c}}_1, ..., \Tilde{\textbf{c}}_4$, their corresponding positions in the original image are computed as:
\begin{equation}
    \Tilde{\textbf{o}}_i = A^{-1} \Tilde{\textbf{c}}_i.
\end{equation}

The face localization prior is then obtained by shading the region bounded by $\Tilde{\textbf{o}}_i$ with value 1, leaving other pixels with value 0.

\begin{figure*}[hbt!]
    \centering
    \includegraphics[width=\linewidth]{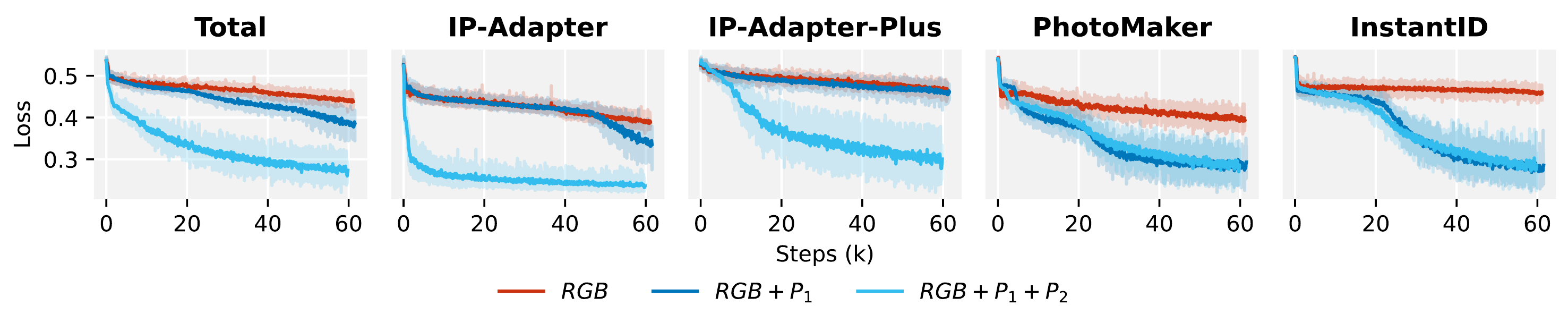}
    \caption{Loss convergence of the first 60k training steps when training with different prior channels. }
    \label{fig:prior_channel_ablation}
\end{figure*}

Our experiments demonstrate that this prior channel is crucial for effective training. As shown by the light and dark blue curves in Fig. \ref{fig:prior_channel_ablation} (corresponding to RGB + $P_1$ in the legend), it significantly reduces both PhotoMaker and InstantID losses. Without this prior, InstantID's cosine similarity plateaus around 0.5 and fails to achieve effective ID protection.

\begin{figure*}[hbt!]
    \centering
    \includegraphics[width=0.65\linewidth]{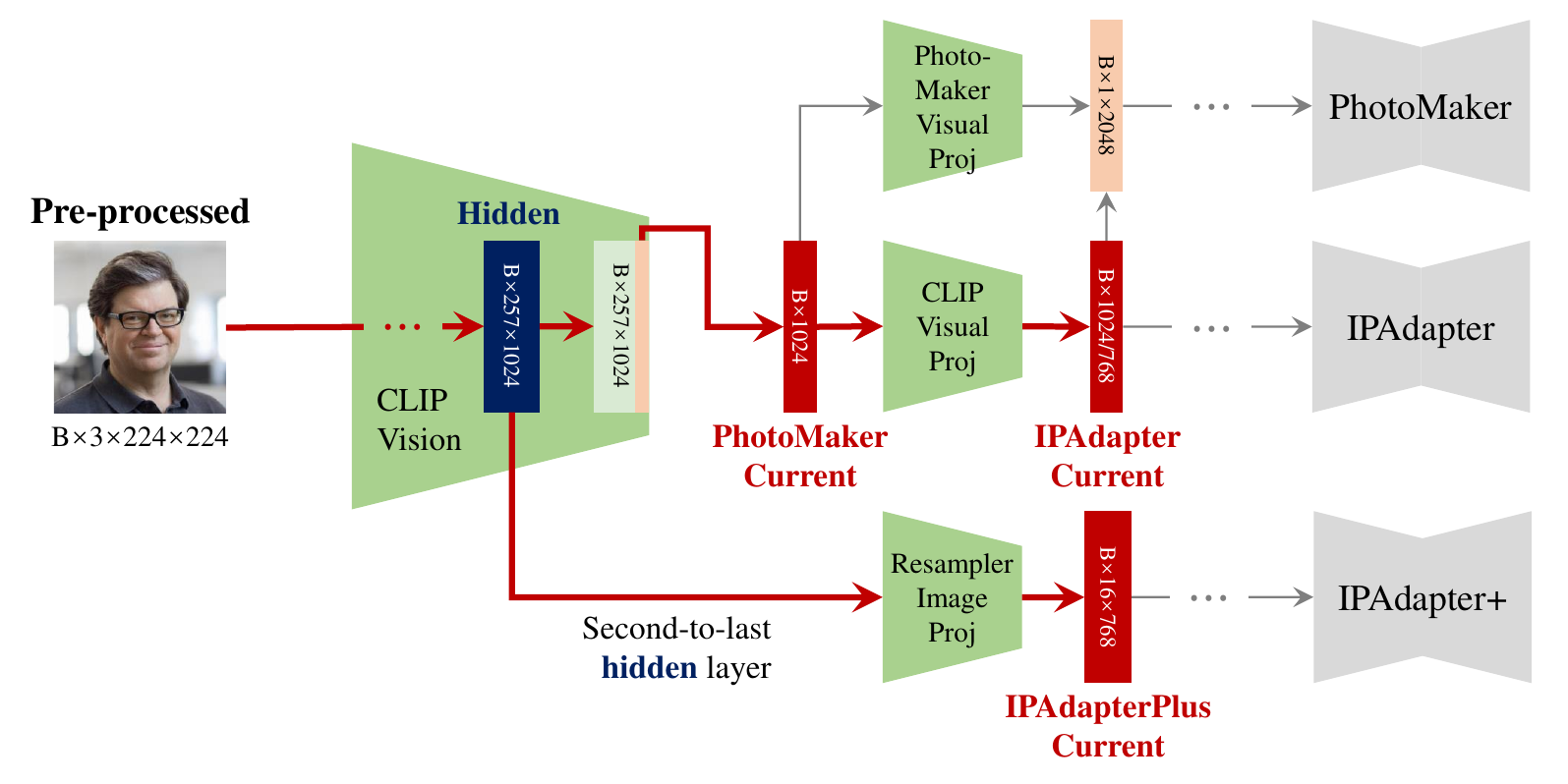}
    \caption{Alternative victim embedding that could be attacked during ID protection.}
    \label{fig:ablation_emb}
\end{figure*}

\subsection{Aspect Ratio Prior}

Similar to the Face Localization Prior, we provide an additional Center Crop Prior channel to the adversarial noise encoder model, serving as its fifth channel and second prior channel input. This channel also consists of binary values (0/1) and indicates the image regions that will be preserved after CLIP preprocessing. Providing this prior effectively communicates CLIP's preprocessing approach to the network through the prior format, eliminating the need for the network to learn this information independently and thereby reducing the training burden. As shown by the light blue curves in Fig. \ref{fig:prior_channel_ablation} (corresponding to RGB + $P_1$ + $P_2$ in the legend), this prior effectively helps IP-Adapter and IP-Adapter-Plus converge more rapidly during the early stages of training.

\section{Ablation on Alternative Target Embeddings}

In Sec. \ref{sec:IP_protection_obj}, we introduced three principles for selecting the target embedding for attacks, including blocking all potential pathways of information flow and ensuring the embedding possesses a certain level of semantic abstraction to facilitate the attack. However, our selection is not the only viable choice.  

Beyond the current option, the penultimate hidden state of the CLIP Vision model also exhibits these two properties to some extent. This embedding is shown in blue in Fig. \ref{fig:ablation_emb}, while the current attack target is marked in red. We investigated replacing the current optimization target with this alternative embedding. The results, shown in Table \ref{tab:ablation_emb}, indicate that while the drop in facial similarity (measured using ISM) is less pronounced compared to the current target, the identity protection performance on IP-Adapter and IP-Adapter-Plus remains nearly comparable. This demonstrates that, although the current embedding achieves superior results, the choice of attack embedding is not uniquely constrained.

\begin{table}[]
\centering
\caption{Protection performance when choosing different target embeddings for attack. Each branch is evaluated using either the second-to-last hidden state (\textcolor{Blue}{hidden}) or the embedding obtained after the CLIP vision output (\textcolor{Red}{current}).}
\label{tab:ablation_emb}
\resizebox{\linewidth}{!}{%
\begin{tabular}{ccccc}
\toprule
\multirow{2}{*}{\textbf{\begin{tabular}[c]{@{}c@{}}Victim\\ Embedding\end{tabular}}} & \textbf{} & \multicolumn{3}{c}{\textbf{ISM $\downarrow$}}                        \\ \cline{3-5} 
                                                                                     & \textbf{} & \textbf{IP-Adapter} & \textbf{IP-Adapter-Plus} & \textbf{PhotoMaker} \\ \hline
\textbf{\textcolor{Blue}{Hidden}}                          &           & 0.047               & 0.075                    & 0.134               \\
\textbf{\textcolor{Red}{Current}}                          &           & 0.042               & 0.074                    & 0.005               \\ \bottomrule
\end{tabular}%
}
\end{table}

\section{More Visualizations on Protecting against Unseen Models}

\begin{figure*}
    \centering
    \includegraphics[width=0.88\textwidth]{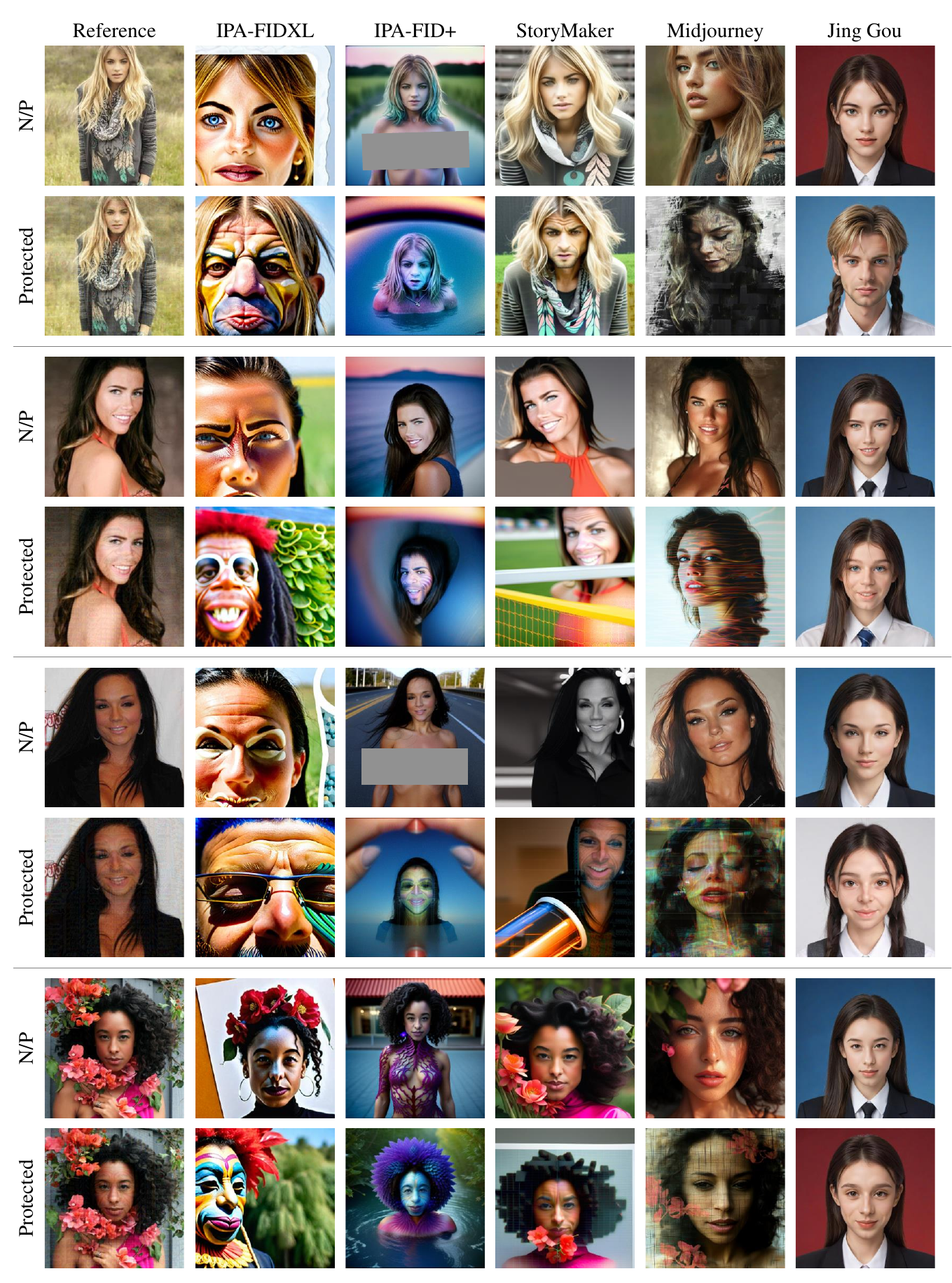}
    \caption{More visualizations of ID protection performance on unseen generators.}
    \label{fig:generalisation_more}
\end{figure*}

Fig. \ref{fig:generalisation_more} provides more visualizations of ID protection performance against customization models unseen during training.

\end{document}